\documentclass[runningheads]{llncs}
\usepackage{hyperref}
\usepackage{graphicx}
\usepackage{booktabs}
\usepackage{csquotes}
\usepackage{xcolor}

\begin{document}
\title{Closing the Loop: Testing ChatGPT to Generate Model Explanations to Improve Human Labelling of Sponsored Content on Social Media}
\titlerunning{Using Model Explanations to Improve Labelling of Sponsored Content}
% %Thales Bertaglia, Stefan Huber, Catalina Goanta, Gerasimos Spanakis and Adriana Iamnitchi
 \author{Thales Bertaglia\inst{1, 3}\orcidID{0000-0003-0897-4005} \and
 Stefan Huber \and
 Catalina Goanta\inst{2}\orcidID{0000-0002-1044-9800} \and
 Gerasimos Spanakis\inst{1}\orcidID{0000-0002-0799-0241} \and
 Adriana Iamnitchi\inst{1}\orcidID{0000-0002-2397-8963}}
% %
\authorrunning{T. Bertaglia et al.}
\institute{Maastricht University, Maastricht, Netherlands \and
Utrecht University, Utrecht, Netherlands \and Studio Europa, Maastricht, Netherlands
}
\maketitle              % typeset the header of the contribution
\begin{abstract}
Regulatory bodies worldwide are intensifying their efforts to ensure transparency in influencer marketing on social media through instruments like the Unfair Commercial Practices Directive (UCPD) in the European Union, or Section 5 of the Federal Trade Commission Act. Yet enforcing these obligations has proven to be highly problematic due to the sheer scale of the influencer market. The task of automatically detecting sponsored content aims to enable the monitoring and enforcement of such regulations at scale. 
Current research in this field primarily frames this problem as a machine learning task, focusing on developing models that achieve high classification performance in detecting ads. These machine learning tasks rely on human data annotation to provide ground truth information. However, agreement between annotators is often low, leading to inconsistent labels that hinder the reliability of models. To improve annotation accuracy and, thus, the detection of sponsored content, we propose using chatGPT to augment the annotation process with phrases identified as relevant features and brief explanations.
Our experiments show that this approach consistently improves inter-annotator agreement and annotation accuracy. Additionally, our survey of user experience in the annotation task indicates that the explanations improve the annotators' confidence and streamline the process. Our proposed methods can ultimately lead to more transparency and alignment with regulatory requirements in sponsored content detection.

%link for FTC Act https://www.federalreserve.gov/boarddocs/supmanual/cch/200806/ftca.pdf
%link for UCPD https://eur-lex.europa.eu/legal-content/EN/TXT/?uri=celex%3A32005L0029 
%link influencer marketing hub https://influencermarketinghub.com/influencer-marketing-benchmark-report/

\keywords{sponsored content detection \and human-AI collaboration \and legal compliance \and social media}
\end{abstract}

\section{Introduction}
The rise of influencers, content creators monetising online content through native advertising, has drastically changed the landscape of advertising on social media~\cite{EUP-influencers,goanta2020regulation}. This shift has increased concern about hidden advertising practices that might harm social media users. For decades, advertising rules have been applied to legacy media in such a way as to separate commercial communication from other types of content. The primary rationale behind rules relating to mandated disclosures has been that hidden advertising leads to consumer deception. Despite the increasing legal certainty that native advertising, such as influencer marketing, must be clearly disclosed, monitoring and enforcing compliance remains a significant challenge~\cite{said2013mandated}.

The task of automatically detecting sponsored content aims to enable the monitoring and enforcement of such regulations at scale. For instance, in the United Kingdom, the Competition and Markets Authority is one of the enforcement agencies tasked with monitoring influencer disclosures on social media, which is done using some automated techniques developed by their internal data unit\footnote{\url{https://www.gov.uk/cma-cases/social-media-endorsements}}. In published scholarship, most existing methods frame the problem as a machine learning task, focusing on developing models with high classification performance. The success of these models depends on the quality and consistency of human-annotated data, which often suffer from low inter-annotator agreement, compromising the reliability and performance of the models~\cite{DBLP:journals/corr/abs-2107-02278,vidgen2020directions}. Moreover, fully-automated approaches are insufficient for regulatory compliance, where human decision-makers are ultimately responsible for imposing fines or pursuing further investigations. 

To bridge this gap, we propose a novel annotation framework that augments the annotation process with AI-generated explanations, which, to our knowledge, is the first attempt in this domain. These explanations, presented as text and tokens or phrases identified as relevant features, aim to improve annotation accuracy and inter-annotator agreement. Our experiments show that our proposed framework consistently increases agreement metrics and annotation accuracy, thus leading to higher data quality and more reliable and accurate models for detecting sponsored content. Critically, our work tackles the need for explainability in AI tools used for regulatory compliance, ensuring that human decision-makers can better understand and trust the outputs of these models. This is particularly important for market surveillance activities, which have not yet caught up with the transparency and accountability issues at the core of discussions around individual surveillance~\cite{kossow2021algorithmic}. 

\section{Related Work}
Sponsored content detection has primarily been studied as a text classification problem. Works in this field generally train models in a semi-supervised setting, using posts disclosed as ads with specific hashtags as weak labels. Generally, there is a lack of focus on evaluating model performance with labelled data. Most works collect their own datasets and do not describe whether (and how) data is annotated. Since social media platforms typically do not allow data sharing, there are no standardised datasets for evaluating the task; thus, comparing results is challenging. Furthermore, the absence of labelled data for evaluation affects the reliability of results, as models are often not tested on undisclosed ads. 

From a technical perspective, previous studies have employed traditional machine learning models with basic text features~\cite{ershovEffectsInfluencerAdvertising2020,waltenrathEmpiricalEvidenceImpact2021a}, neural networks with text embeddings~\cite{zareiCharacterisingDetectingSponsored2020}, and multimodal deep learning architectures combining text, image, and network features~\cite{kimDiscoveringUndisclosedPaid2021b,kimMultimodalPostAttentive2020}. In this paper we experiment with some of these models in addition to chatGPT and GPT-4 for classification. Although peer-reviewed research is limited due to chatGPT's recent release, some technical reports have found chatGPT to achieve state-of-the-art performance in several text classification tasks~\cite{pikuliak2023chatgpt,wei2023chainofthought,gilardi2023chatgpt}.

Interdisciplinary research combining computational methods with fields such as communication and media studies and law has focused on identifying influencers, describing their characteristics, and mapping the prevalence of their disclosures~\cite{mathurEndorsementsSocialMedia2018,arriagadaYouNeedLeast2020,christinDramaMetricsStatus2021}. In the context of using explanations to improve data labelling or decision-making, research has explored AI-human collaboration and investigated the optimal integration of explanations for human interaction~\cite{Kim_2023,neerincx2018using,van2019pluggable,van2020allocation}. To the best of our knowledge, our paper is the first to propose using AI-generated explanations to improve the detection of sponsored content, bridging the gap between explainable AI and regulatory compliance in the context of sponsored content on social media.

\section{Experimental Setup}
This section describes the dataset we use, how we selected the model for sponsored content detection, generated explanations to augment the annotation process, and designed the annotation task and the user-experience survey.

\subsection{Data Collection}
We collected and curated our own dataset of Instagram posts for this study. We manually selected 100 influencers based in the United States using the influencer discovery platform Heepsy\footnote{\url{https://heepsy.com}}. We selected 50 micro-influencers (between $100k$ and $600k$ followers) and 50 mega-influencers (over $600k$ followers). Then, we collected all available data and metadata from all posts for each account using CrowdTangle\footnote{\url{https://www.crowdtangle.com/}}, the Meta platform that provides access to social media data for (among others) academic purposes. Our dataset includes $294.6k$ posts, 66.1\% from mega-influencers and 33.9\% from micro-influencers. CrowdTangle's Terms of Service do not allow (re)sharing datasets that include user-generated content; thus, we cannot share the full dataset. However, the list of the \textit{ids} of accounts and posts is publicly available on~\url{https://github.com/thalesbertaglia/chatgpt-explanations-sponsored-content/}

\subsection{Detecting Sponsored Content}
In the first step of our experimental setup, we aim to select the most suitable sponsored content classifier for generating explanations. We evaluate three previously proposed models%chosen based on related work and their performance in other text classification tasks. These models are
: (1)~a logistic regression classifier with term frequency inverse-document frequency (TF-IDF) features, analogous to the approach used by~\cite{ershovEffectsInfluencerAdvertising2020,waltenrathEmpiricalEvidenceImpact2021a}, (2)~a pre-trained BERT model fine-tuned for our task, comparable to~\cite{kimDiscoveringUndisclosedPaid2021b,zareiCharacterisingDetectingSponsored2020}, and (3)~OpenAI's chatGPT (GPT-3.5-turbo as of March 2022), which achieves state-of-the-art results in various text classification tasks~\cite{gilardi2023chatgpt,pikuliak2023chatgpt}. We generate GPT predictions using OpenAI's API.

To evaluate the models' performance, we select a sample from our original dataset and split the data into training and test sets by year, using 2022 for testing and all prior posts for training. This division simulates a real-world scenario where a model is deployed and used to classify unseen data for regulatory compliance. By ensuring no temporal overlap between the sets, we prevent the model from learning features correlated with a specific period. Given the high imbalance in the data (only 1.72\% of posts are disclosed as sponsored), we apply the random undersampling approach proposed by Zarei et al. (2020)~\cite{zareiCharacterisingDetectingSponsored2020} to balance the data. We include all disclosed posts ($n$) and randomly sample ($2*n$) posts without disclosures as negative examples. We allocate 90\% of the balanced data before 2022 to training and the remaining 10\% to validation. We use all data in 2022 as the test set.

Additionally, we labelled a sample of the test set to evaluate the model's performance in detecting undisclosed ads. Four annotators labelled 1283 posts in total, with a sample of 50 posts labelled by all annotators for calculating agreement metrics. The inter-annotator agreement was 52\% in absolute agreement and 53.37 in $\alpha$, indicating moderate agreement. 654 posts were labelled as sponsored (50.97\%) and 629 as non-sponsored (49.03\%). 91.59\% of the sponsored posts did not have disclosures -- i.e., they were identified as undisclosed ads.

We employ a semi-supervised approach to train the models, treating disclosed sponsored posts as positive labels for the \textit{sponsored} class. We consider \textit{\#ad}, \textit{\#advertisement}, \textit{\#spons}, and \textit{\#sponsored} as ad disclosures. We then remove disclosures from the posts to prevent models from learning a direct mapping between disclosure and sponsorship. We train the logistic regression model using TF-IDF features extracted from word-level n-grams from the captions (unigrams, bigrams, and trigrams). For the BERT-based model, we use the \textit{bert-base-multilingual-uncased} pre-trained model weights from HuggingFace~\cite{wolf-etal-2020-transformers}. We fine-tuned the BERT-based model for three epochs using the default hyperparameters (specified in Devlin et al. (2019)~\cite{devlin-etal-2019-bert}).

We apply various prompt-engineering techniques to enhance GPT's predictions. As we use the same methodology for generating explanations, we provide a detailed description in the following subsection. We evaluate all models using F1 for the positive and negative classes, Macro F1 (the simple average of both classes) and Accuracy in detecting undisclosed ads -- a critical metric for determining the models' effectiveness in detecting sponsored posts without explicit disclosures, which is ultimately our goal. \autoref{tab:clf_results} presents the classification metrics for the three models, calculated based on the labelled test set.

\begin{table}
\centering
\caption{Performance of the different models on the labelled test set. Acc represents the models' accuracy in detecting undisclosed ads.}
\label{tab:clf_results}
\begin{tabular}{lcccc}
\toprule
 \bfseries Model & \bfseries Pos F1  & \bfseries Neg F1 & \bfseries Macro F1  & \bfseries Acc \\
\midrule
\bfseries Log Reg & 45.33 & 66.50 & 55.92 & 28.71 \\
\bfseries BERT & 29.30 & 68.84 & 49.07 & 10.85 \\
\bfseries GPT-3.5 & 76.09 & 63.93 & 70.01 & 88.98 \\
\bottomrule
\end{tabular}
\end{table}

GPT-3.5 outperforms the other models in Macro F1 and accuracy in detecting undisclosed ads. Logistic regression (Log Reg) and BERT achieve significantly low accuracy, suggesting their inability to identify undisclosed sponsored posts effectively. The difference in Macro F1 is smaller, highlighting that relying solely on this metric for evaluating models may not accurately reflect their actual performance. Therefore, having high-quality labelled data, including undisclosed ads, is crucial for proper evaluation.

BERT's inferior performance compared to Log Reg could be due to a few factors. Being pre-trained on longer texts, BERT might struggle to extract sufficient contextual information from short Instagram captions. In addition, Log Reg, when combined with TF-IDF features, effectively captures word-level n-grams that may be more effective at identifying sponsored content patterns. In contrast, BERT uses subword tokenisation, which could result in less efficient pattern recognition. Given GPT-3.5's superior performance, particularly in detecting undisclosed sponsored posts, we selected it as the model for generating explanations to augment the annotation task.

% In preliminary tests, we found that GPT-3.5 was generally performant in detecting hints for sponsored content and GPT-4 even more so, but we also acknowledge that AI-based classifcations are of limited use for law enforcement for regulatory reasons. However feeding human decision makers only with posts GPT prefiltered as 'likely sponsored' and augmenting the human decision makers with a) the decision relevant words/parts of the post and b) a reasoning of why GPT argues it may contain undisclosed ads could improve both speed and quality of human decision makers. Consequently, we generate a list of relevant words from the posts using GPT and produce explanations of why or why not they are likely the result of a paid partnership.

\subsection{Generating Explanations with GPT}
\label{sec:generating_explanations}
We investigated various prompts for all publicly accessible models from the GPT-3 series and GPT-4. We observed that even the smallest GPT-3 model, Ada (\textit{text-ada-001}), performed well in sponsored content detection and identifying relevant words. Nevertheless, we noted significant performance improvements for larger models especially when employing chain-of-thought reasoning~\cite{wei2023chainofthought} and generating explanations -- particularly for more ambiguous posts. Consequently, we focused on \textit{GPT-3.5-turbo} (the default ChatGPT version as of March 2022) and GPT-4.

We found a conservative bias for both models, with a strong preference for predicting the \textit{not sponsored} class or other negative labels over positive ones. This phenomenon appeared consistent across all \textit{Davinci-} and \textit{Curie-}based models, with the inverse being true for smaller \textit{Babbage} and \textit{Ada-}based models. We employed several prompt engineering techniques to mitigate this bias and calibrate the labels. First, we instructed the model to highlight relevant words and generate explanations before classifying a post. This chain-of-thought prompting approach, inspired by~\cite{wei2023chainofthought}, significantly reduced bias and improved prediction interpretability. Second, we used few-shot learning to refine explanation calibration, address known failure modes, and further alleviate bias~\cite{NEURIPS2020_1457c0d6}. Third, we experimented with different label phrasings, such as \enquote{Likely (not) sponsored}, to enhance the model's ability to make less confident predictions. Finally, we directly instructed the model to favour positive labels in cases of uncertainty, aiming to identify a higher proportion of undisclosed ads. The final prompt is available on the project's GitHub repository~\footnote{\url{https://github.com/thalesbertaglia/chatgpt-explanations-sponsored-content/}}.

Upon qualitative evaluation, we found that GPT-4 outperformed GPT-3.5-turbo in explanation quality and classification accuracy, especially for ambiguous posts. However, for this study, we chose GPT-3.5-turbo (hereafter referred to as \enquote{GPT}) due to its advantages in speed, cost, and public accessibility. Following this approach, we obtained the most important words in a post and generated explanations for why a post may or may not be sponsored to assist annotators. The following is an illustrative example of such an explanation; we omitted the actual brand name to ensure the post's anonymity:
\begin{verbatim}
Key indicators: '@BRAND', 'LTK'.
The post promotes a fashion brand and features a discount code,
indicating a partnership. Additionally, it features a @shop.LTK
link, a platform for paid partnerships.
\end{verbatim}

\subsection{Annotation Task}
\label{sec:annotation_task}
We conducted a user study to evaluate how explanations can help detect sponsored content. The study consisted of an annotation task in which participants labelled 200 Instagram posts from our dataset as \textit{Sponsored} or \textit{Non-Sponsored}. Our objective with the task was two-fold: i) Analyse explanations as a tool for improving annotation as a resource for ML tasks -- i.e., to measure their impact on data quality, which, in turn, allows for the development of better models and evaluation methods. ii) Simulate regulatory compliance with sponsored content disclosure regulations -- i.e., how a decision-maker would flag posts as sponsored.

We framed the annotation as a text classification task in which annotators had to determine whether an Instagram post was sponsored based on its caption. Generally, we followed the data annotation pipeline proposed by Hovy and Lavid (2010)~\cite{hovy2010towards}. We instructed annotators to consider a post as sponsored if the influencer who posted it was, directly or indirectly, promoting products or services for which they received any form of benefits in return. These benefits included direct financial compensation and non-monetary benefits, such as free products or services. Self-promotion was an exception: we considered posts promoting the influencer's content (e.g. YouTube channel or podcast) non-sponsored. However, posts advertising merchandise with their brand or directly selling other goods still fall under sponsored content. We explained these guidelines to each annotator and provided examples of sponsored and non-sponsored posts to help reinforce the definitions.

Eleven volunteer annotators with varying levels of expertise participated in the study. All were between 20 and 30 years old, active social media users, and familiar with influencer marketing practices on Instagram. Additionally, all annotators had or were working towards a high-education degree in a European university. Demographically, the participants came from various countries. We did not specifically collect country-level information, but at a continent level, participants were from Asia, Europe, and South America. While all participants were fluent in English, none were native speakers. 

We split annotators into three groups according to their level of expertise in annotating sponsored content on social media. The first group, with three people, consisted of participants with no prior experience in data annotation. The second group included four participants who previously participated in annotation tasks but had no formal training. The third group, consisting of four legal experts, had specific legal expertise in social media advertisement regulations and had participated in annotations before. We further split the subgroups of annotators into two groups regarding annotation setup: one without explanations, in which annotators only had access to the captions, and one augmented with the generated explanations. One group of four annotators labelled the posts in both setups: with and without explanations. To summarise, our study includes three distinctive groups: novices with no prior annotation experience, intermediate annotators with previous experience but no formal training, and legal experts knowledgeable in social media regulations.

To select the 200 Instagram posts for our user study, we turned to a sample previously labelled by law students in another annotation task. Although the labels and definitions used in that task differed from ours, they provided a way to identify which posts were undisclosed ads, allowing us to include them in our study. We selected posts published between 2017 and 2020 by 66 different influencers based in the United States, with 62\% being mega-influencers and 38\% being micro-influencers. We also included 15\% of posts with clear ad disclosures (such as the hashtag \#ad) as an attention check to ensure annotators noticed the disclosures. Based on the labels from the previous annotations, we estimate that 65\% (130) of the posts were likely sponsored, and 50\% (100) were likely undisclosed ads.

We set up the study using the open-source annotation platform Doccano\footnote{\url{https://github.com/doccano/doccano}}. Each participant had a unique project, and although all annotators labelled the same 200 posts, the labels were not shared, and each participant only had access to their annotations. The annotation interface displayed the caption of the post and the two possible labels (Sponsored and Non-Sponsored) as buttons. After the post caption, we added the generated explanations with an explicit delimitation. 

Accurately measuring inter-annotator agreement is crucial in data annotation tasks, as it allows us to estimate the annotated data's quality and the decision-making process's reliability. To assess inter-annotator agreement in our study, we used three main metrics: Krippendorff's Alpha ($\alpha$), absolute agreement, and accuracy in detecting disclosed posts. Krippendorff's Alpha measures the degree of agreement among annotators, considering the level of agreement expected by chance alone~\cite{krippendorff2011computing,hayes2007answering}. The absolute agreement indicates the proportion of annotations where all annotators agreed on the same label. We also used accuracy in detecting disclosed posts as an attention check mechanism, as it measures annotators' ability to correctly identify posts with clear disclosures as sponsored. This metric is crucial because disclosures may not always be easily visible in posts~\cite{mathurEndorsementsSocialMedia2018}. We also analysed additional metrics in some experiments, which we will introduce when describing the specific experiments.

\subsection{User-experience Survey}
After the annotation, we conducted a user-experience survey to gather feedback from annotators on their experience using the explanations to assist with their decision-making process. The survey consisted of seven questions, with five closed-ended and two open-ended questions. We describe all questions and the rating scale used below:

\begin{itemize}
    \item \enquote{On a scale of 1 (not helpful) to 5 (extremely helpful), how helpful were the explanations in identifying undisclosed advertisement partnerships?}
    \item \enquote{How accurate, from 1 (extremely inaccurate) to 5 (extremely accurate), did you think the explanations were?}
    \item \enquote{How often, from 1 (0\% of the time) to 5 (100\% of the time), did you agree with the AI explanations?}
    \item \enquote{Did the AI explanations help you feel more confident in your decision-making (Yes/No)?}
    \item  \enquote{What aspects of the AI explanations were most helpful for your decision-making process?} This was a multiple-choice question with five options: \textit{Reasoning}, \textit{Identifying specific words or phrases}, \textit{Clear examples}, \textit{Other (specify)}, and \textit{None}.
    \item \enquote{In what ways did the AI explanations improve your understanding of what constitutes an undisclosed advertisement partnership?} Open-ended.
    \item \enquote{How could the AI explanations be further improved to better support your decision-making process? Did you find anything noticeable you want us to know?} Open-ended.
\end{itemize}

The participants who received annotations augmented with explanations all completed the questionnaires, and we ensured their anonymity by not collecting any identifiable information. Additionally, we made it clear to the annotators that their responses would be entirely anonymous.

\section{Experimental Results}
This section presents the main findings from the annotation task and user-experience survey. \autoref{tab:basic_agreement} shows the metrics comparing the agreement between annotators who labelled the posts with and without explanations. Seven participants were in the \textbf{No Explanations} group (one with \textit{no experience}, four with \textit{some experience}, and two \textit{legal experts}). The \textbf{With Explanations} group had eight people (three with \textit{no experience}, three with \textit{some experience}, and two \textit{legal experts}) -- one participant from the \textit{no experience} group and three from \textit{some experience} labelled in both settings. In addition to the metrics presented in~\autoref{sec:annotation_task}, we also evaluate the proportion of posts with at most one disagreement (\textit{1-Disag}) and show the percentage of posts labelled as sponsored (\textit{Sponsored}). The last two rows present the absolute and relative (normalised) differences in metrics between the groups. The relative differences in metrics indicate the proportional change (in percentage). Positive differences represent an increase in agreement. 

\begin{table}
\caption{Agreement metrics comparing annotations with and without explanations.}
\label{tab:basic_agreement}
\centering
\begin{tabular}{lccccc}
\toprule
 & \bfseries $\alpha$ & \bfseries Abs & \bfseries 1-Disag & \bfseries Acc & \bfseries Sponsored \\
\midrule
\bfseries No Explanations & 54.98 & 46.50 & 69.50 & 90.62 & 54.64 \\
\bfseries With Explanations & 63.58 & 54.50 & 75.00 & 93.75 & 59.81 \\ \midrule
\bfseries Absolute Diff & 8.61 & 8.00 & 5.50 & 3.12 & 5.17 \\
\bfseries Relative Diff & 15.65 & 17.20 & 7.91 & 3.45 & 9.46 \\
\bottomrule
\end{tabular}
\end{table}

Using explanations to enhance the annotations resulted in a consistent improvement across all inter-annotator agreement metrics. Specifically, there was a 15.65\% increase in $\alpha$ and a 17.20\% increase in absolute agreement. However, the final values were still relatively low, typical of annotations in complex decision-making tasks~\cite{DBLP:journals/corr/abs-2107-02278,geiger2020garbage,vidgen2020directions}. Accuracy in detecting disclosed posts also improved by 3.45\%, but the final result was not perfect, suggesting that annotators still fail to identify all disclosure hashtags, even with explanations highlighting them. Additionally, the proportion of posts labelled as sponsored increased by 9.46\%, indicating that explanations led annotators to identify more as sponsored. We also analyse the agreement between all pairs of annotators to measure the variation in agreement and ensure the reliability of the annotations. \autoref{tab:basic_pairwise_agreement} summarises the pairwise agreement metrics. The \textit{Min} and \textit{Max} columns represent the lowest and highest agreement metric values among the annotator pairs, respectively, and the $\pm$ column denotes the standard deviation.

\begin{table}
\centering
\caption{Pairwise agreement comparing annotations with and without explanations.}
\label{tab:basic_pairwise_agreement}
\begin{tabular}{lcccccc}
\toprule
 & \bfseries Min Abs & \bfseries Max Abs & \bfseries $\pm$ & \bfseries Min $\alpha$ & \bfseries Max $\alpha$ & \bfseries $\pm$ \\
\midrule
\bfseries No Explanations & 66.00 & 88.50 & 5.28 & 30.81 & 77.04 & 10.83 \\
\bfseries With Explanations & 73.00 & 90.00 & 4.49 & 43.13 & 79.53 & 10.00 \\ \midrule
\bfseries Absolute Diff & 7.00 & 1.50 & -0.79 & 12.31 & 2.48 & -0.82 \\
\bfseries Relative Diff & 10.61 & 1.69 & -14.98 & 39.96 & 3.22 & -7.62 \\
\bottomrule
\end{tabular}
\end{table}

The pairwise metrics reveal considerable variation in the agreement between annotator pairs. For the \textit{No Explanation} group, there was a substantial difference of 46.23 in $\alpha$ between the pair with the lowest and highest agreement, with a standard deviation of 10.83. This difference indicates that some annotators are significantly less reliable than others. However, the group \textit{With Explanations} showed a consistent improvement, with less variation between pairs. The standard deviation decreased by 14.98\% for absolute agreement and 7.62\% for $\alpha$, indicating more reliable annotations. Even the lowest-agreement pair showed significant improvement, with an increase of 10.61\% for absolute agreement and 39.96\% for $\alpha$. These results suggest that using explanations to augment annotations led to a higher inter-annotator agreement overall, improved consistency between pairs, and even increased agreement among the least reliable annotators. To better understand the impact of augmenting the annotation with explanations, we also investigated how it affects different subgroups of annotators. We divided the subgroups into three categories: legal experts, non-experts, and annotators who labelled in both settings (with and without explanations) -- this category does not include legal experts. \autoref{tab:subgroup_agreement} presents the agreement metrics for each category in both subgroups of annotators, as well as the relative difference between them. \# indicates the number of participants within the subgroup. For clarity, we did not report the proportion of annotations with at most one disagreement because some subgroups contain a single pair of annotators.

\begin{table}
\centering
\caption{Agreement metrics for different subgroups of annotators, aggregated according to their expertise level.}
\label{tab:subgroup_agreement}
\begin{tabular}{lccccc}
\toprule
 & \bfseries $\alpha$ & \bfseries Abs & \bfseries Acc & \bfseries Sponsored & \bfseries \# \\
\midrule
\bfseries Legal Experts No Explanations & 52.11 & 76.50 & 96.88 & 57.25 & 2 \\
\bfseries Legal Experts With Explanations & 61.94 & 83.00 & 100.00 & 66.50 & 2 \\ 
\bfseries Relative Diff & 18.86 & 8.50 & 3.23 & 16.16 & - \\ \midrule
\bfseries Non-Experts No Explanations & 62.04 & 62.50 & 93.75 & 53.60 & 5 \\
\bfseries Non-Experts With Explanations & 64.89 & 59.50 & 93.75 & 57.58 & 6 \\
\bfseries Relative Diff & 4.59 & -4.80 & 0.00 & 7.43 & - \\ \midrule
\bfseries Labelled Both No Explanations & 66.74 & 70.00 & 96.88 & 53.12 & 4 \\
\bfseries Labelled Both With Explanations & 73.15 & 74.50 & 100.00 & 54.50 & 4 \\
\bfseries Relative Diff & 9.60 & 6.43 & 3.23 & 2.59 & - \\
\bottomrule
\end{tabular}
\end{table}

The annotations augmented with explanations showed consistent improvements in all subgroups, except for absolute agreement within the non-expert group. Legal experts had the most significant improvement in $\alpha$ (18.86\%). Additionally, the proportion of posts labelled as sponsored increased significantly (16.16\%), with the subgroup \textit{Legal Experts With Explanations} having the highest value (66.5\%). This subgroup and \textit{Labelled Both With Explanations} achieved 100\% accuracy in detecting disclosed sponsored posts. \textit{Labelled Both} also had the highest $\alpha$ in both settings. It is important to note that higher agreement does not necessarily imply higher accuracy in correctly identifying sponsored posts. The metrics measure how much a subgroup of annotators agree on the definitions they are applying to label; they could be wrongly applying a consistent judgement. Therefore, we cannot reliably conclude which group had the best performance. Moreover, the high agreement within the subgroup \textit{Labelled Both} could be influenced by the annotators labelling the same posts twice in both settings. Although we randomly shuffled the posts to reduce the likelihood of memorisation, repetition could still affect agreement. Nevertheless, the high proportion of sponsored content and absolute agreement for the annotation within \textit{Legal Experts With Explanations} indicate that experts agree that there are more sponsored posts than non-experts tend to identify.

While explanations can improve the quality of annotations, they may also introduce bias by influencing annotators to rely on specific cues presented in the explanation; annotator bias is a common challenge in text annotation tasks~\cite{alkuwatly-etal-2020-identifying,geva2019modeling}. To investigate potential bias introduced by explanations in our study, we examine whether annotators tended to use the same label predicted by GPT. Although we did not explicitly provide GPT's prediction as part of the explanation, the model's reasoning and highlighted words and phrases might imply the predicted label, leading to over-reliance on the model and decreasing the accuracy of annotations. Thus, it is essential to analyse the impact of GPT's predictions on annotator behaviour to ensure the reliability and fairness of the annotations. Specifically, we calculate two metrics -- the distribution of posts labelled as sponsored and the majority agreement with GPT predictions -- to compare the agreement between annotators who received explanations and those who did not. We use majority agreement instead of absolute to reduce the impact of low-agreement pairs and fairly compare all groups. If the agreement with GPT predictions increased in the group with explanations, it could indicate that annotators followed the model's predictions. We hypothesise that, for the \textit{Labelled Both} group, an increase in agreement with GPT predictions proportionally more than the percentage of sponsored posts would suggest that annotators changed their judgements based on the model's cues. \autoref{tab:gpt_bias} summarises the results of this analysis.

\begin{table}
\centering
\caption{Proportion of posts labelled as sponsored and majority agreement with GPT predictions across subgroups of annotators.}
\label{tab:gpt_bias}
\begin{tabular}{lcc}
\toprule
 & \bfseries Sponsored & \bfseries Agreement \\
\midrule
\bfseries No Explanations & 54.64 & 85.50 \\
\bfseries With Explanations & 59.81 & 90.50 \\ 
\bfseries Relative Diff & 9.46 & 5.85 \\ \midrule
\bfseries Legal Experts No Explanations & 57.25 & 77.50 \\
\bfseries Legal Experts With Explanations & 66.50 & 92.00 \\
\bfseries Relative Diff & 16.16 & 18.71 \\ \midrule
\bfseries Non-Experts No Explanations & 53.60 & 81.00 \\
\bfseries Non-Experts With Explanations & 57.58 & 88.50 \\
\bfseries Relative Diff & 7.43 & 9.26 \\ \midrule
\bfseries Labelled Both No Explanations & 53.12 & 78.50 \\
\bfseries Labelled Both With Explanations & 54.50 & 87.00 \\
\bfseries Relative Diff & 2.59 & 10.83 \\
\bottomrule
\end{tabular}
\end{table}

The majority agreement with GPT predictions is consistently high across all subgroups, ranging from 77.5\% to 92\%. All subgroups that received explanations had an increase in agreement with GPT predictions compared to the corresponding No Explanations subgroup. Specifically, except for \textit{Labelled Both}, all subgroups showed proportional increases in both metrics, indicating no clear bias for GPT predictions. However, the \textit{Labelled Both} subgroup demonstrated a significant increase in agreement with GPT predictions compared to the proportion of sponsored posts, suggesting that the annotators changed their decision-making process after having access to explanations. While this result indicates a bias towards the model's predictions, more experiments are needed to determine its impact on data quality. Given the generally high accuracy of GPT demonstrated in our classification experiments, relying on them could improve annotation accuracy.

On the other hand, the difference in agreement with the predictions between the \textit{Legal Experts} subgroups adds uncertainty about the model's accuracy. The subgroup of legal experts with no explanations had the lowest agreement with GPT predictions; in contrast, those with explanations had the highest. The groups include different annotators, and \textit{Legal Experts No Explanations} had low inter-annotator agreement; therefore, we cannot effectively measure the model's accuracy. Although we found evidence of explanations biasing the annotators, further research is needed to investigate how this result impacts data quality. 

Finally, we conducted a user-experience survey to gather feedback from annotators on their experience using the explanations to assist with their annotation process. All the responses are available online on \url{https://tinyurl.com/sponsored-annotation-survey}. We ensured that the document preserves the anonymity of all parties involved in the study.

The survey results showed that 87.5\% of annotators felt more confident in their decision-making with the help of explanations. Additionally, 62.5\% rated the explanations highly helpful and accurate (4 out of 5). Only one participant rated them as unhelpful (2 out of 5). The average estimate of agreement with the explanations was close to the agreement with GPT predictions, with 62.5\% of annotators estimating that they agreed with the explanations between 80\% and 100\% of the time. Notably, all annotators selected the words and phrases highlighted by the model explanations as a helpful feature, while only 37.5\% selected the reasoning behind the predictions. This result indicates a preference for precise explanations. Comparable explanations could be generated from any classifier using local-explainability methods such as LIME~\cite{lime}. This shows that the methodology proposed and evaluated in our study does not rely on GPT's capability of generating longer text-based explanations and could be reproduced with simpler models.

The open-ended questions revealed two clear trends among participants. First, most participants found the highlighted words and phrases helpful in identifying brands and context-relevant hashtags in the posts. Second, participants suggested that adding the likelihood of a post being sponsored as a feature would be a useful improvement to the explanations. Overall, these results indicate that participants had a positive experience with the explanations, found them helpful and accurate, and felt they improved their decision-making.

\section{Summary}
Our experiments show that inter-annotator agreement metrics consistently improve when augmenting the annotation process with explanations. We observed a 15.65\% increase in $\alpha$ and a 17.20\% increase in absolute agreement among the general population of annotators. The accuracy in detecting disclosed sponsored posts improved by 3.45\%, and the proportion of posts labelled as sponsored increased by 9.46\%. These findings indicate that explanations not only help annotators identify more sponsored content but also enhance the reliability of annotations and reduce variation between annotator pairs. Our user-experience survey shows that most annotators found the explanations helpful and accurate, increasing their trust in decision-making. Therefore, our proposed annotation framework could lead to higher-quality data labelling and improve decision-makers' experience in regulatory compliance contexts. We made the\textit{ids} of posts in our dataset, along with all the labels annotated by annotators and the GPT predictions, publicly available~\footnote{\url{https://github.com/thalesbertaglia/chatgpt-explanations-sponsored-content/}}, offering a valuable resource that could benefit research in the field.

Nevertheless, our study has some limitations. One potential issue is the bias introduced by explanations, as annotators may rely on specific cues presented in the explanation. While we found no clear bias for most subgroups, we note that the group that labelled posts in both settings showed a significant increase in agreement with GPT predictions compared to the proportion of sponsored posts. Another area for improvement is the small sample size of legal experts and the variation in agreement metrics among different subgroups, which may impact the generalisability of our results.

Future research should investigate the impact of explanations on annotator bias and data quality and explore open-source models with greater transparency, such as LLaMA~\cite{touvron2023llama}, instead of OpenAI's GPT -- which is a privately-owned model with limited information regarding its training data. Moreover, conducting experiments with larger and more diverse samples of annotators, including more legal experts, could shed light on the role of expertise in the annotation process. Expanding the study to other annotation tasks and domains would also provide insights into the generalisability of our findings, potentially benefiting a broader range of applications.

Despite these limitations, it is important to consider that digital enforcement and market monitoring by authorities such as consumer agencies will exponentially grow in the coming years. Thus, monitoring techniques must consider transparency and explainability to avoid accuracy issues when applying legal sanctions.  

\bibliographystyle{splncs04}
\bibliography{main}




% LNCS Homepage, \url{http://www.springer.com/lncs}. Last accessed 4
% \end{thebibliography}
\end{document}